\documentclass{article}

\usepackage{arxiv}

\usepackage[utf8]{inputenc} % allow utf-8 input
\usepackage[T1]{fontenc}    % use 8-bit T1 fonts
\usepackage{hyperref}       % hyperlinks
\usepackage{url}            % simple URL typesetting
\usepackage{booktabs}       % professional-quality tables
\usepackage{amsfonts}       % blackboard math symbols
\usepackage{nicefrac}       % compact symbols for 1/2, etc.
\usepackage{microtype}      % microtypography
\usepackage{lipsum}

\usepackage[bottom]{footmisc}

\usepackage{setspace} 
\usepackage[utf8]{inputenc} %unicode support
\usepackage{amsmath, amsthm, amssymb}	
\usepackage{amssymb}               % Voor speciale symbolen zoals de verzameling Z, R...
\usepackage{graphicx}              % Om figuren te kunnen verwerken
\usepackage{epstopdf}
\usepackage{commath}
\usepackage{lscape}
\usepackage{amsfonts}
\usepackage{icomma} %geen spatie achter de komma
\usepackage[small,bf,hang]{caption}

\usepackage{algorithm}
\usepackage[noend]{algpseudocode}

\usepackage{booktabs}
\usepackage{pgfplots}
\usepackage{pgfplotstable}

\author{
	Yanou Ramon\thanks{Corresponding author.} \\
	Department of Engineering Management\\
	University of Antwerp\\
	Antwerp, Belgium\\
	\And
	Sandra C. Matz \\
	Department of Management\\
	Columbia Business School\\ New York City, United States\\
	\And
	R.A. Farrokhnia\\
	Columbia Business \& Engineering Schools\\
	New York City, United States\\
	\And
	David Martens \\
	Department of Engineering Management\\
	University of Antwerp\\
	Antwerp, Belgium\\
}

\begin{document}

\title{Explainable AI for Psychological Profiling from Digital Footprints: A Case Study of Big Five Personality Predictions from Spending Data}
\maketitle


\begin{abstract}
Every step we take in the digital world leaves behind a record of our behavior; a digital footprint. Research has suggested that algorithms can translate these digital footprints into accurate estimates of psychological characteristics, including personality traits, mental health or intelligence. The mechanisms by which AI generates these insights, however, often remain opaque. In this paper, we show how Explainable AI (XAI) can help domain experts and data subjects validate, question, and improve models that classify psychological traits from digital footprints. We elaborate on two popular XAI methods (rule extraction and counterfactual explanations) in the context of Big Five personality predictions (traits and facets) from financial transactions data ($N$ = $6,408$). First, we demonstrate how global rule extraction sheds light on the spending patterns identified by the model as most predictive for personality, and discuss how these rules can be used to explain, validate, and improve the model. Second, we implement local rule extraction to show that individuals are assigned to personality classes because of their unique financial behavior, and that there exists a positive link between the model's prediction confidence and the number of features that contributed to the prediction. Our experiments highlight the importance of both global and local XAI methods. By better understanding how predictive models work in general as well as how they derive an outcome for a particular person, XAI promotes accountability in a world in which AI impacts the lives of billions of people around the world.
\end{abstract}

% Introduction
\section{Introduction}
\label{Introduction}
\begin{figure}[ht]
	\centering
	\scalebox{0.75}{\includegraphics{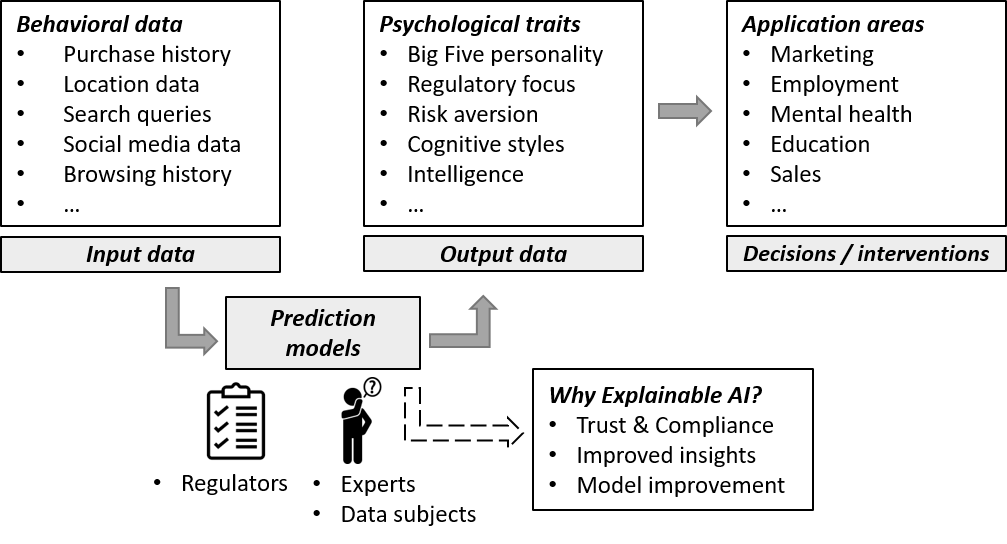}}
	\caption{Explainable AI in applications that leverage behavioral data for psychological profiling.}
	\label{fig:XAI_Psychologicalprofiling}
\end{figure}

The information age is characterized by a wealth of user-generated data that is collected with every step a user takes in the digital environment. These digital footprints are increasingly available for academics, businesses and governments \cite{MatzNetzer} and have been shown to provide highly intimate insights into people's lives as well as the ways in which they think, feel and behave. For example, digital footprints can be used to predict personality traits~\cite{Kosinski2013,MatzPersonalization}, mental health \cite{Moshe2021}, sexual and political orientation~\cite{Kosinski2013,PraetPolitics} or intelligence \cite{Kosinski2013}. The process of translating digital footprints into meaningful psychological profiles with the help of machine learning has been termed 'psychological profiling', and drives applications in a variety of areas ranging from marketing to employment to mental health (see \textbf{Fig. \ref{fig:XAI_Psychologicalprofiling}} for a conceptual overview). As \cite{MatzPrivacy} define it, psychological profiling is ``the automated assessment of psychological traits from digital footprints.'' Over the past decade, researchers have been tapping into a broad variety of data sources for psychological profiling, including social media data (e.g., Facebook likes and status updates~\cite{Kosinski2013,Youyou2015}), mobile sensing data~\cite{deMontjoye2013}, music listening preferences~\cite{Rentfrow2003,Nave2018}, mobility behaviors~\cite{MullerPeters}, as well as financial transaction records~\cite{SpendingPersonality,TovanichSpending}.

% Figure

\subsection{AI as a black box}
Machine learning models that classify psychological traits from behavior can be highly accurate. However, at the same time, their structure can be very complex, which has earned them the reputation of being a `black box' that is difficult to penetrate. The complexity arises from either the learning technique (e.g., Random Forest models), the data, or both. Consequently, it is often difficult---if not impossible---to understand how classifications were made when using nonlinear models without relying on interpretation techniques like the ones we use in this study. Even for linear models or decision trees, it can be challenging to gain meaningful insights into how classifications are made, because of the high dimensional and sparse nature of behavioral data \cite{MartensProvostEDC,RamonCounterfactuals,RamonMetafeatures,ClarkProvost}. For example, if we want to predict people's personality based on the Facebook pages they `like', a user is represented by a binary feature for every page, which results in an enormous feature space. Linear models trained on these data end up having a large number of features (i.e., every Facebook page becomes a separate feature in the model), each of which is assigned a corresponding weight. Alternatively, only the features with the largest weights can be inspected. Because the data is sparse, however, only a small fraction of the classified instances is `explained'. \cite{Kosinski2013}, for example, predicted personal traits using over 50,000 Facebook pages and interpreted the models by listing the pages that are most related to a trait of interest. Amongst the top predictors for high intelligence were pages like `Science' and `Curly Fries'. Because of the data sparsity, however, these pages are only relevant (`liked') by a small fraction of all users predicted as intelligent, which leaves a substantial part of the classifications unexplained (on average, a user liked $170$ pages out of a total of 55,814 pages that were used by the model). 

In addition to the outlined challenges associated with the high dimensionality and sparsity of digital footprint data, the non-redundancy of the data also impacts the ability to meaningfully interpret model predictions. Given that many behavioral features are relevant for the classification task, applying feature input selection or dimensionality reduction generally results in worse predictive performance, and makes a detailed interpretation of the model impossible \cite{DeCnuddeBenchmark, JunqueBiggerBetter, ClarkProvost}. %[add refs] 
Taken together, the high dimensionality and sparsity of digital footprint data in combination with the explosion in potentially relevant features, drive the complexity of models developed from behavioral data, making them difficult to interpret. 

\subsection{Why the interpretability of AI matters}
The lack of transparency and inability to explain decisions of AI systems for psychological profiling from digital traces creates challenges for their adoption. We distinguish three main reasons for the need of interpretability: (1) trust and compliance, (2) insights, and (3) model improvement. 

\subsubsection{Trust and Compliance}
Explaining model outputs helps validate and justify the relations learned from the data and compare this with theoretical assumptions and domain knowledge. This can increase trust of experts to eventually accept these systems \cite{PwC2019,MartensEthics}. Trusting a model implies believing its reliability or truth \cite{MartensEthics}. Next to the model's out-of-sample predictive performance, the explanation needs to provide evidence that the model learned a meaningful pattern that is not only useful in specific circumstances. The need for trust and validation also stems from increasing regulatory pressure. Both the United States (US) and European Union (EU) are pushing toward a regulatory framework for transparent and accountable AI, and global organizations like OECD and G20 aim for a more human-centric approach. Especially for systems deemed as high-risk (defined in the EU's recent AI Act, and referring to every system that can negatively impact the life of a human), explainability has emerged as a key business and regulatory challenge. For example, systems that regulate access to financial services, educational opportunities or employment fall in this category. Psychological profiling can also be part of such applications. Think of talent acquisition and management systems that assess job-relevant characteristics\footnote{In the last decade, many companies have been created that leverage AI for more fair, efficient and effective talent acquisition and management, for example, advertising online job vacancies or measuring the fit of job seekers with open roles in a company using behavioral data (game-based assessments or video interviews). Examples are pymetrics (https://www.pymetrics.ai/) and Humantic AI (https://humantic.ai/).}, or systems to prioritize medical aid (e.g., to people who display early signs of depression \cite{Moshe2021,MatzNetzer}).

Appropriate human-machine interface tools should be put in place that allow experts to interpret the model outputs and overrule them when necessary. This is also important to guarantee safe and fair AI systems that do not exhibit differential effects on subgroups or underrepresented groups \cite{Stachl2020}, which can open up organizations to legal entanglements or cause reputational damage \cite{Amazon2018,WachterFinancialTimes}. In HR analytics, for example, when predicting which persons to invite for an interview, based on resumes and behavioral assessments, it is important to know why a model makes decisions, to ensure there is no unfair treatment of certain groups like women or immigrants (for example, think of the algorithmic discrimination in Amazon's male-biased hiring tool \cite{Amazon2018} or Uber running job ads targeted exclusively at men \cite{WachterFinancialTimes}). Interpretability techniques might not directly solve these issues, but can be used as a tool to audit models and detect sources of bias that can arise from skewed data collection or real human bias hidden in the data. 
In addition, regulatory requirements and increasing customer expectations push companies to provide transparency to those affected by the data-driven decisions (hereafter `data subjects'). For example, the General Data Protection Regulation (GDPR) notes the `right to explanation' for those affected by decisions of AI systems.

\subsubsection{Improved insights} 
A second reason for model explainability stems from a broader goal of predictive modeling: to learn something about a domain. Interpretability allows researchers and domain experts to verify knowledge encoded in the models, that can be useful for building on prior research, or for theory building and exploratory work. For example, businesses might ask: what are the main reasons we are inviting job applicants to an interview based on their resumes and motivations? Psychologists might wonder: what are the behavioral manifestations of people on social media who suffer from burn-outs? Explaining AI systems helps explore new insights, that, in turn, can inspire new hypotheses to be tested with more traditional, statistical methods. Insights into personality predictions from behavior can also lay groundwork for (research on) interventions targeting specific behavior (e.g., to promote well-being \cite{MullerPeters}).

In addition, improved insights might translate into a competitive advantage when companies are able to \textit{share} these insights with consumers.  In targeted marketing and sales efforts relying on the prediction of psychological profiles \cite{MatzNetzer}, for example, explainability could be used to validate the models predictions and meet the needs of demanding customers who want both control and service \cite{MatzPrivacy,ChenCloaking}.
As \cite{MatzPrivacy} argue, insight into not just the data that is collected but also the inferences that are derived from it, can help data consumers make more informed decisions that are based on trade-offs between improved service and privacy. In line with this, non-profit initiatives like mePrism\footnote{https://www.meprism.com/} and Digita\footnote{https://www.digita.ai/} aim to give insight to online users on the data that's collected about them and how companies use this information, and support them to be in charge over their digital footprints. The European Commission's Digital Services Act (DSA) further emphasizes this by noting that recipients of online advertisements should get ``meaningful explanations of the logic used'' for ``determining that specific advertisement is to be displayed to them'' (paragraph 52). Another example of giving insights to data subjects is providing personalized feedback to job candidates on data-driven insights about their strengths, development needs, and organizational fit, that can in turn guide them in future job search endeavors. Moreover, this can improve the candidate experience and the overall quality of the recruiting process, and eventually benefit the company as well \cite{HBR_HR}.

\subsubsection{Model improvement}
Explanations can be used to improve prediction models and identify weaknesses that arise from models overfitting to the data and/or perpetuating historical biases. When modeling human behavior, monitoring the important predictors of a model is crucial, for example, to identify reasons for drops in performance over time, that can be caused by changing behavior; a phenomenon known as `concept drift' (for example changing spending behavior in times of a pandemic \cite{Farrokhnia}; we refer to \cite{Stachl2020} and \cite{Lu2018} for more examples).
Technology and culture are evolving at a rapid pace which means that the purpose of technical devices and the way we interact with them are constantly changing. The information captured by online behavior can thus change over time and lead the model's performance to drop \cite{Lu2018}. Although a number of control mechanisms can be put into place (e.g., online learning \cite{Lu2018,Mittal2015}), understanding which behavioral features have a (large) impact on a model's classifications through explanations can help domain experts make sound statements on the expected lifetime of a model and its sensitivity to rapidly changing technological indicators and digital behavior. For example, the type of mobile phone applications that people use might change more rapidly compared to the genres of movies people watch or the type of places they visit on the weekend, which reflect more `stable' behavior.

Explanations can also help understand the generalization ability of a model beyond the sample data or improve overall troubleshooting. For example, some behavior can be specific to populations located in geographical regions. If the model picks up these specific behaviors and gives them a large weight in the decisions, this might limit the usability of that model in other contexts.

\subsection{Using Explainable AI to overcome black box approaches: Research Overview}
Over the last decade, a growing body of  research has been dedicated to the field of Explainable Artificial Intelligence (XAI). The aim of this research area is to develop and apply algorithms to explain prediction models and individual predictions. The desire to have both predictive and interpretable models resulted in an explosion of new methods to extract useful information from black box models.\footnote{A detailed overview of all XAI methods proposed in the literature is beyond the scope of this study, so we point readers interested in learning more about all different techniques to recent overviews, for example, \cite{XAIreview}, \cite{2018Guidotti}, or \cite{Molnar}. In this study, we select methods that we believe are particularly suitable to explain classifications of models from large-scale behavioral data.}

In this study, we build on this research to address the challenge of interpretability at a global and local level: We apply techniques that provide global insight into why a model makes classifications of interest (e.g., when does the model typically classify someone as Neurotic based on their behavior?) and more granular, local explanations for why a particular decision was made (e.g., why does the model classify this person as Neurotic?). 
The main contribution of this study lies in demonstrating the use and importance of XAI methods to interpret classifications about people's psychological traits from behavioral data, be it for model acceptance, validation, insights, or improvement. We are the first to apply rule extraction and counterfactual explanations, two popular XAI techniques, in the field of psychological profiling. 

The remainder of this paper is structured as follows:  In Section \ref{RelatedWork}, we describe the XAI techniques and motivate why we select these methods in this paper. Next, in Section \ref{MaterialsMethods}, we describe the data and methods used in our case study. We apply XAI in the context of Big Five personality predictions from real-world financial transactions data collected by a non-profit organization in the United States ($N$ = $6,408$). To bring in an angle that goes beyond the prediction task in related work, we model personality hierarchically: we model both traits and their underlying facets (e.g., Extraversion can be broken down in facets: Assertivism, Energy, and Sociability). In Section \ref{Results}, we first discuss the classification performance of the models, and then go over the observations from the model interpretability analysis. In Section \ref{Discussion}, we summarize the main findings and their implications, and point at a good deal of room for further research at the intersection of XAI and computational psychology. Finally, Section \ref{Conclusions} sets out the conclusions of this study.

\section{Introduction to the field of Explainable AI (XAI)}
\label{RelatedWork}
As described in Section \ref{Introduction}, models that classify psychological traits from behavioral data are often considered `black box' approaches. That is, it is generally difficult to determine why and under which conditions a class of interest (hereafter also referred to as the `positive class') was predicted. In an attempt to open the black box, the field of XAI field has started to develop tools and frameworks that provide insights into how models work, providing human experts with the ability to understand the logic that goes into the algorithm's decisions. %[ref Google]. 
A large body of work has focused on post hoc explanations to extract information about a model's behavior without addressing details of their inner workings. Instead, these methods only use the input data and the model's predicted outputs. One of the most prominent advantages of post hoc explainability is that interpretations can be provided after developing complex models without needing to sacrifice predictive performance \cite{Molnar}.

Explanation methods can have a global or local scope. Global explanations give insight into models at an aggregate level, over all the model's classifications. Local explanations explain individual classifications. In this study, we use both \textit{global} and \textit{local} XAI methods to explain \textit{classifications}\footnote{There exists a subtle yet important distinction between explanation methods that explain (discrete) classifications vs. (continuous) predicted scores (we refer to \cite{Fernandez} for a full discussion).
In this study, we focus on explaining classifications that drive concrete decisions and/or actions to be taken.} of a model $C_{M}$, that predict a psychological trait $\textbf{Y}$ (\textit{i.e., target variable}) from behavioral data $X$ $\subset$ $\mathbb{R}^{N \times M}$, where $N$ and $M$ respectively indicate the number of data subjects (\textit{i.e., instances}) and features. Note that we solely focus on classification tasks in this study. 
Although prior work on psychological profiling has predominantly focused on predictive performance \cite{Settani,Stachl2020}, there have been attempts at explaining the underlying mechanisms.  Some studies have highlighted the face validity of predictive models by showing the most related predictors to the target (based on univariate correlations prior to modeling that not necessarily reflect what the model learned from the data \cite{Kosinski2013}), or by providing a list of important features \cite{TovanichSpending,deMontjoye2013,Stachl2019}. While such approaches offer initial insights, they do not reflect how the (combination of) feature values impact(s) the predicted classes, nor the extent to which the classifications are explained. Understanding the latter is particularly valuable when modeling very sparse data where one feature might only be relevant to a small number of instances (e.g. liking `Curly Fries' on Facebook might be predictive of IQ, but only a small fraction of the population likes `Curly Fries' on Facebook). In this paper, we therefore move beyond what has previously been suggested by the literature. Our selection of methods is based on the following criteria: We exclude methods that might not be suitable when modeling high-dimensional behavioral data. For example, visualizations of feature effects are mentioned in \cite{Stachl2020} as a way to increase interpretability in personality computing applications, by tracing how the outcome variable changes as the value of a feature changes (e.g., score on Extraversion). However, we argue that this approach is not appropriate for models with hundreds to thousands of features, where many features might be relevant for the task, and for which the important features may vary substantially between classifications (as we will demonstrate in Section \ref{Results}). Users who want to understand how a specific variable relates to the predicted outcome (either at an aggregate or local level) might still benefit from using this approach, however, it is impractical to show how classifications come about by showing the effect of just one or two features (i.e., interaction plot).

As a \textit{global} XAI method, we therefore use rule extraction to capture under which conditions a class of interest is predicted, and discuss how these explanation rules can be used to validate learned relations, generate new hypotheses, and identify weaknesses of the model. To explain predictions at the \textit{local} level, we use counterfactual explanations that reveal which features contributed to a single classification, or more precisely, point to changes of the feature values that lead the model to make another decision. In the following subsections, we go over rule extraction and counterfactual explanations in more detail.

\subsection{Rules as global explanations}
We use rule extraction as a global method to gain insight into the classification models. Rule extraction has been proposed in the literature to generate explanations by distilling a comprehensible set of rules (hereafter `explanation rules') from a complex classification model $C_{M}$. Rule extraction is based on surrogate modeling of which the goal is to use an interpretable model to approximate the predictions of a more complex model $\hat{\textbf{Y}}$. The interpretable model used as surrogate can be a concise set of if-then-else rules (in which case it's called `rule extraction') or a linear model with a small number of features. The complexity of the rules is restricted so that the final explanations are comprehensible to humans.\footnote{Rule extraction can be challenging for high-dimensional, sparse data, as the black box model needs to be replaced by many rules to explain a substantial fraction of the classifications, which leaves the user again with an incomprehensible explanation \cite{RamonMetafeatures}. To address this, \cite{RamonMetafeatures} proposed a technique based on metafeatures (i.e., clusters of the original features) to extract a concise set of rules that more accurately approximates the model's behavior. In this study, however, we apply rule extraction on the original data, because the dimensionality and sparsity of the data used in the case study are still manageable.} A main motivation for the use of rule extraction is to combine the desirable predictive behavior of complex classification techniques with the comprehensibility of decision trees and/or rules.

We use rule extraction for different reasons. First, an important advantage of rule extraction is that the learned relations between features and predicted classes are not lost. Another advantage is that it approximates \textit{classification} behavior of a model. This is in contrast to other XAI methods, like feature relevance methods, that do not reflect how features impact predicted classes, but merely provide a list of important features \cite{Fernandez}. Moreover, using rule extraction---or surrogate models in general---we can quantify the extent to which the model is explained using a metric called $Fidelity$.\footnote{If we use a linear model as surrogate to approximate the model's behavior, we can also compute $Fidelity$ of the explanation. However, the same limitations as for feature relevance lists hold. The information about the interaction between features and their correspondence to the class gets lost. Moreover, it gets more difficult to grasp the classification behavior. There will exist a very large number of conditions that explain when the model predicts a particular output, rendering the explanation less comprehensible.} $Fidelity$ can be operationalized in different ways. Here, we refer to the metric that computes the overlap between the predicted classes of the model $\hat{\textbf{Y}}$ and the classes predicted by the explanation rules $\hat{\textbf{Y}}_{rules}$ as $Fidelity$.\footnote{Essentially, you can compare $Fidelity$ to $Accuracy$ that is used as a performance metric in a traditional machine learning context. \textit{Accuracy} measures to what extent the model's predictions $\hat{\textbf{Y}}$ overlap with the ground-truth classes $\textbf{Y}$. In contrast, \textit{Fidelity} measures to what extent the explanation rules' predicted classes $\hat{\textbf{Y}}_{rules}$ overlap with the model's predicted classes $\hat{\textbf{Y}}$.} The goal is to extract rules that have high $Fidelity$, i.e., approximate the patterns learned in the original model to the best possible extent. For imbalanced problems, it is often more insightful to use the $Fscore$ of predicting the output of the model to measure the quality of the explanation, which we refer to as \textbf{$Fscore_{f}$}. $Fscore_{f}$
is measured by the harmonic mean between $Recall_{f}$ and $Precision_{f}$, and reflects how well the `positive class' is explained by the rules. $Recall_{f}$ measures the proportion of positives predicted by the model that are retrieved, and $Precision_{f}$ measures the proportion of correct classifications among the instances predicted as a class of interest (a `positive') by the rules. All else equal, we prefer an explanation rule set that results in a higher \textbf{$Fscore_{f}$}, because this explanation reflects the original model's predictions more accurately. We measure the quality of rules on an out-of-sample test set, as we want the explanation to reflect the model's prediction behavior on \textit{new} data, not just on the training data.\footnote{The same challenges of overfitting in machine learning hold in the surrogate modeling context. As an extreme example, consider a decision table as an explanation that memorizes when the model predicts a class of interest. For new data, the table would never classify someone as a class of interest (the persons' identifiers will never match an identifier in the table). We would get a high in-sample, but a low out-of-sample $Fidelity$, because the decision table does not reflect how the model is actually making classifications from the data.}

\subsection{Counterfactual rules as local explanations}
For explaining model classifications at the local level, we compute counterfactual rules \cite{MartensProvostEDC,RamonCounterfactuals,Wachter}. Compared to local feature relevance methods, such as Local Interpretable Model-agnostic Explanations (LIME) \cite{LIME} and SHapley Additive exPlanations (SHAP) \cite{SHAP}, counterfactual rules explain the model's predicted class instead of the score \cite{RamonCounterfactuals,Fernandez}. Following Martens \& Provost (2014), who defined counterfactuals for document classifications, we define counterfactual rules for a classification as a set of features from the instance that is causal: changing the value of the features causes the system's decision to change. In other words, the decision would have been different if not for the presence of this set of features. 
There are multiple ways of defining changes of the feature values. A common approach is to simulate the `missingness' of a feature by replacing the value by the mean value of the feature (for continuous features) or the mode value (for sparse numerical, binary or categorical data). In essence, we are asking ourselves the question if the model would make the same decision if a feature in question would be missing \cite{Fernandez}. It is important, both in research and practice, that the choice on how to define `changes' is clearly mentioned, because, depending on this, slightly different explanations may arise.

We use counterfactual explanations, first of all, because they point at a set of features without which the AI system would have made a different decision. They help us understand how features affect \textit{decisions} of AI systems, rather than predicted scores, in terms of domain knowledge, rather than in terms of modeling techniques. Providing a concrete justification for a decision gives data subjects insight into changes to receive a desired result in the future, based on their current behavior, and is consistent with requirements specified in regulatory frameworks \cite{Wachter}. Another advantage is that the explanation only comprises a (small) fraction of all features used in a model, which makes it a particularly interesting approach to explain decisions of models with high-dimensional feature dimensions. Prior work showed cases where these explanations can be obtained only in seconds for models on large-scale data, and that explanations typically consisted of a handful to a few dozen of features \cite{MartensProvostEDC,ChenCloaking,RamonCounterfactuals}. Moreover, in contrast to local feature relevance methods, where it is non-trivial to choose the complexity setting (i.e., how many features to show), the answer for counterfactuals is clear-cut: those features are shown that allow for the creation of a counterfactual rule \cite{RamonCounterfactuals}.

\section{Case Study: Predicting personality traits from financial transaction records}
\label{MaterialsMethods}
We use a case study on the prediction of Big Five personality traits from real-world transactions data to demonstrate how global and local XAI methods can help shed light on the ways by which the prediction model learns and makes decisions about the target individual. 

\begin{figure}[ht]
	\centering
	\scalebox{0.7}{\includegraphics{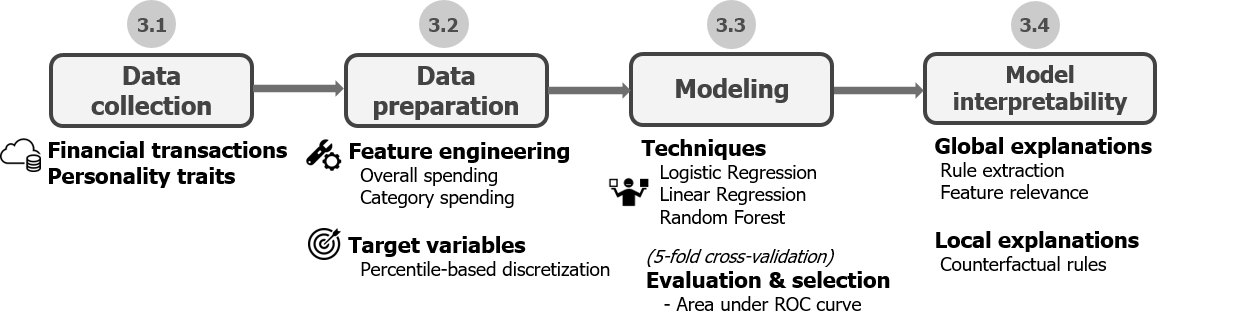}}
	\caption{Methodology of the case study to develop models that classify people's personality from financial transactions ($3.1$ to $3.3$) and gain insight into the final models by means of global and local XAI methods ($3.4$).}
	\label{fig:MLworkflow}
\end{figure}

\textbf{Fig. \ref{fig:MLworkflow}} depicts the methodology used in our case study. We describe (i) how the data was collected ($3.1$), (ii) how the data was prepared for the analyses ($3.2$), (iii) the model specifications ($3.2$), as well as (iv) the ways in which XAI can help understand and validate the models ($3.4$).\footnote{This methodology can be applied more generally to psychological profiling applications that mine other types of behavioral data, such as social media data, GPS location data, and web browsing histories.} In what follows, we go over each step in more detail. In Section 3.5, the results of the case study are discussed.

% Figure

\subsection{Data collection}

\subsubsection{Financial transactions}
We use financial transactions data collected by a non-profit based in the US. The organization offers a money management application to help people manage their savings more effectively. Individuals can join the platform by linking their bank accounts, including checking, savings, and credit card accounts. Using these data, the organization provides people with financial decision-making aid and motivates them to achieve savings objectives by offering rewards and lotteries. As part of their onboarding experience, users can voluntarily complete a personality questionnaire. For the purpose of our case study, we use de-identified historical transactions between January and December $2019$. We subset the data to active accounts to guarantee a sufficient amount of data per person: we discard individuals with fewer than five transactions or less than \$$100$ spent on a monthly basis, or fewer than five distinct spending categories.\footnote{The users of the money management application have relatively large financial constraints. For this reason, we set more flexible criteria compared to related studies (see for example \cite{TovanichSpending})% and \cite{Gotz})
.} This exclusion procedure leaves us with $N$ = $6,408$ data subjects of whom we have transactions data (linked to their self-reported personality profiles) that can be fed into the prediction models. The transactions data include a time stamp indicating when the transaction was made, the amount of the transaction (in US Dollar), and the category of the transaction. Each transaction belongs to one of the $285$ spending categories.

\textbf{Table~\ref{table:descriptives}} shows summary statistics of the sample data. The individuals observed in the data generally have a low-income profile, i.e., they spend, on average, about \$$15,000$ over the course of one year. In the US, a household has on average \$$63,036$ of expenditures per year~\cite{USBureauLaborStatistics}. With an average household consisting of $2.5$ adults, this is equivalent to annual expenditures of \$$25,214$ per capita. The average amount of yearly expenditures for low-income households with a total income before taxes less than \$$15,000$ is lower, and equals \$$15,745$ per capita \cite{USBureauLaborStatistics}.

% Table
\begin{table}
	\centering
	\caption{Summary statistics of the transactions data. The data contains $4,539,634$ spending records between January and December 2019 of $6,408$ data subjects. Only individuals with at least five transactions and \$$100$ spent in each month, and at least five distinct spending categories are selected. There are $285$ spending categories and the average household consists of $3$ people.}
	\label{table:descriptives}
	\scalebox{0.85}{
	\begin{tabular}{c|c|c} %c|c}
	\noalign{\smallskip}\hline
	\textbf{Per customer} & \textbf{Mean} (\textbf{Std}) & \textbf{Median}\\
			\noalign{\smallskip}\hline
			\\
			Total amount transactions & \$$47,236.26$ (\$$58,441.34$) & \$$33,649.78$\\
			\\
			Amount per transaction & \$$66.77$ (\$$53.94$) & \$$16.71$\\
			\\
			Number of transactions & $708.43$ ($441.49$) & $621$\\
			\\
			Unique number of spending categories & $43.66$ ($16.12$) & $43$\\
			\\
			\noalign{\smallskip}\hline
			\textbf{Per spending category} & \textbf{Mean} (\textbf{Std}) & \textbf{Median}\\
			\noalign{\smallskip}\hline
            \\
			Total amount transactions & \$$1,062,070$ (\$$4,142,055$) & \$$29,256.64$\\
			Rel. total amount transactions & $0.0035$ ($0.014$) & $9.7\text{e-}5$\\
            \\
			Number of transactions & $15,928.54$ ($51,812.71$) & $544$\\
			Rel. number of transactions & $0.0035$ ($0.011$) & $1.2\text{e-}4$\\
			\\
			Customer support & $981.79$ ($1,494.70$) & $240$\\
			Rel. customer support & $0.15$ ($0.23$) & $0.04$\\
	\end{tabular}}
\end{table}

\subsubsection{Personality traits}
Personality traits are conceptualized as relatively stable characteristics that explain and predict differences in cognition, affect and behavior. Decades of research have suggested that there are five dimensions that explain these individual differences across a broad variety of contexts, including different cultures or language. These five dimensions are known as the Big Five (BF) Model of Personality~\cite{CostaMcCrae1992}. The BF model proposes five traits that capture individual differences in the way people think, feel and behave~\cite{CostaMcCrae1992}: (1) \textit{Extraversion}, the tendency to seek stimulation in the company of others, to be outgoing and energetic; (2) \textit{Agreeableness}, the tendency to be warm, compassionate, and cooperative; (3) \textit{Conscientiousness}, the tendency to show self-discipline, aim for achievement, and be organized; (4) \textit{Neuroticism}, the tendency to experience unpleasant emotions easily; and (5) \textit{Openness to Experience} (or simply \textit{Openness}), the tendency to be intellectually curious, creative, and open to feelings~\cite{CostaMcCrae1992,TovanichSpending}.
Personality theory specifies that traits are hierarchically organized~\cite{CostaMcCrae1992,BFI2S}: each domain subsumes more specific facets that have a unique variance not entirely explained by the higher order Big Five.\footnote{Adaptations of the original Big Five Inventory (BFI) questionnaire (e.g., BFI-$2$-S)---that was not intended as hierarchical measure---allow to simultaneously assess someone's personality at the trait and facet level.} The facets vary slightly across models and measures, but for the purpose of this case study we leverage  the Big Five Inventory (BFI-2) questionnaire which suggests the following facets:
\begin{itemize}
    \item Extraversion: Sociability, Assertiveness, Energy
    \item Agreeableness: Compassion, Respectfulness, Trust
    \item Conscientiousness: Organization, Productivity, Responsibility
    \item Neuroticism: Anxiety, Depression, Emotional Volatility
    \item Openness: Intellectual Curiosity, Aesthetic Sensitivity, Creative Imagination
\end{itemize}

\begin{figure}[h!]
	\centering
	\scalebox{0.5}{\includegraphics{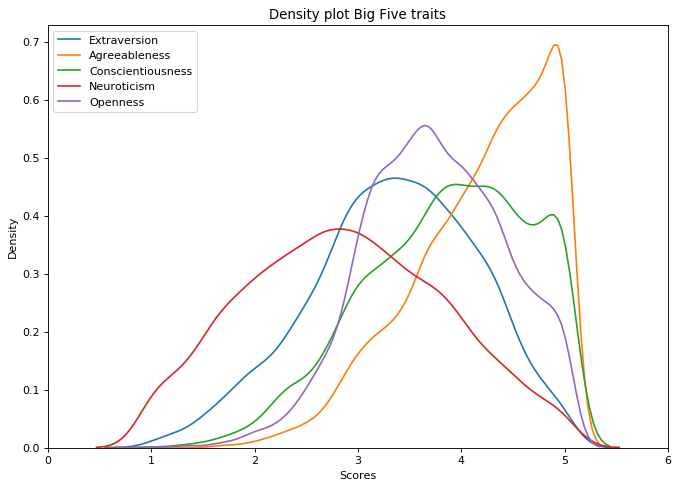}}
	\caption{Distribution of scores on the Big Five traits in our sample of $6,408$ individuals.}
	\label{fig:DensityBFI2Domains}
\end{figure}

Our sample data contains the (self-reported) BF personality traits of the data subjects at the trait ($5$) and facet ($15$) level. All $6,408$ individuals completed a personality survey and provided their consent to have their transactions history matched with their survey responses for the purpose of this study. The traits were measured by the established BFI-$2$-S questionnaire~\cite{BFI2S}, in which participants indicate their agreement with $30$ statements using a five-point Likert scale ($1$=`Disagree strongly' to $5$=`Agree strongly'). Each trait (resp., facet) was measured using a six-item (resp., two-item) scale and the final (averaged) scores range between $1$ and $5$, respectively indicating a low or high score. With Cronbach's alpha being larger than  $.7$ across all Big Five traits (Extraversion=$.80$, Agreeableness=$.79$, Conscientiousness=$.82$, Neuroticism=$.85$, Openness=$.72$), internal consistencies were found to be good. \textbf{Fig.~\ref{fig:DensityBFI2Domains}} shows the distribution of the traits. Neuroticism follows a normal distribution, whereas Extraversion and Openness are approximately normally distributed. The distributions of Conscientiousness and Agreeableness are skewed to the left, indicating that the majority of individuals in the sample perceive themselves as highly agreeable and conscientious. \textbf{Table A\ref{table:Internet_sample}} shows the mean and standard deviation of the traits in the sample under investigation and compares this against a reference sample of $1,000$ American individuals (i.e., the Internet sample in Soto \& John~\cite{BFI2S}). 

% *** Fig. ***

\subsection{Data preparation}

\subsubsection{Feature engineering}
The spending data reflect a wide range of behavioral patterns, which we organize into two broad categories of features based on related work \cite{TovanichSpending,SpendingPersonality}: (1) \textit{overall spending} that comprises summary statistics of spending aggregated over time and features that enrich the aggregated measures with finer-grained, time-dependent information (e.g., how much does the daily amount someone spends vary over time?) and (2) \textit{category spending} that reflects a person's spending category profile and relative spending per category. In total, we extract $578$ features from the raw transactions data. Calculations and definitions of the features are detailed in \textbf{Table A\ref{table:FeatureEngineering}}.

\textit{Overall spending.} For every individual, we compute the total number of transactions $n_{tot}$ and the total amount someone spent $a_{tot}$ aggregated over the $12$-month period. We also compute the average amount spent per transaction $a_{avg}$ and the (relative) variability of the transaction amount $a_{cv}$ defined as the ratio between the standard deviation and the mean of the transaction amount.\footnote{We use the coefficient of variation because it is a more robust measure when comparing the variance of two variables with different means (i.e., the average amount of money spent per transaction varies between individuals).} A low variability indicates that a person spends money equally over different transactions. Lastly, we measure the average daily amount spent $a_{avg,daily}$ and the (relative) variability of the daily transaction amount $a_{cv,daily}$ which is computed similar to $a_{cv}$ but then on a daily basis. A low value for $a_{cv,daily}$ indicates that someone spends their money equally over different days.

\textit{Category spending.} For every individual, we compute their spending proportions in each category: we calculate both the relative amount of transactions $n_{c}$ and the relative amount of money $a_{c}$ that a person spent in each category $c$. Their transactions are mapped to the $285$ spending categories, then aggregated and normalized to get the percentage of spending in each category. We also compute the number of unique spending categories $C_{tot}$ and the diversity of spending over different categories $C_{entropy}$. A high value of $C_{entropy}$ indicates that someone equally distributed their transactions over the spending categories in which they made transactions. A low value indicates that a person has transactions that are distributed over a few categories.

\subsubsection{Target variables}
Each person in the data is characterized by a set of historical financial transactions and a (self-reported) score for each of the traits. Following prior work \cite{deMontjoye2013,TovanichSpending,Stachl2020}, we define a multi-class classification task for each of the traits by splitting the data into three classes (High vs. Middle vs. Low), where we create discrete classes in the continuous scale scores using a percentile-based approach~\cite{deMontjoye2013,TovanichSpending}.\footnote{Classes of personality can also be constructed using a central tendency estimate \cite{Pianesi2008}, however, this can result in a high rate of misclassifications. Big Five traits tend to be normally distributed \cite{Phan2021,Stachl2020}, which means that many scores lie close to the central tendency estimate of the scale (see \textbf{Fig.~\ref{fig:DensityBFI2Domains}}). Consequently, the artificial `Low vs. High' distinction results in a greater separation between subjects than actually exists. Further, this approach likely results in a large number of misclassifications due to measurement error, i.e., the true scores on BF traits of each individual may be close, but not exactly equal to, the measured values.} We specifically focus on the High and Low classes. This decision was driven by the fact that the higher and lower classes are often those of interest in applied contexts, where it is useful, for example, to know which individuals are highly extraverted and therefore have certain behavioral tendencies.  For example, companies might want to adjust their marketing message to the outgoing and social nature of extraverts or select the most conscientious candidates for a job interview.

We use min-max normalization to transform the raw scores into a decimal between $0$ and $1$. The normalized scores are used to develop the regression models (e.g., Logistic Regression), which can in turn be used to make classifications using a threshold (an approach known as regression-based classification). Second, we use percentile-based discretization to map the scores to personality buckets. To construct a binary target that indicates if someone scores High on a trait, we transform the scores that exceed the $66$th percentile to $1$, else $0$. In a similar fashion, we construct another binary variable that indicates if someone scores Low on a trait using the $33$rd percentile.

\subsection{Modeling}

\subsubsection{Modeling techniques} 
Machine learning algorithms can be used to make classifications of psychological traits about new individuals. These algorithms are suitable for large-scale data, such as behavioral data, and allow to model complex relationships. Moreover, the algorithms can pick up on subtle patterns of which humans are unaware or cannot perceive~\cite{Stachl2020}. We test both linear and nonlinear models from the data and select the final classification model using a five fold cross-validation procedure to test out-of-sample performance. For linear models, we train regularized Linear and Logistic Regression models.\footnote{We use both lasso and ridge regularization.} We also train Random Forest models that account for possible nonlinearities between the behavioral features and the target personality classes. Random Forest classifiers (resp., regressors) are ensemble learners that fit a number of decision tree classifiers (resp., regressors) on various subsamples of the data and use averaging to improve the out-of-sample accuracy.

\begin{figure}[h!]
	\centering
	\scalebox{0.65}{\includegraphics{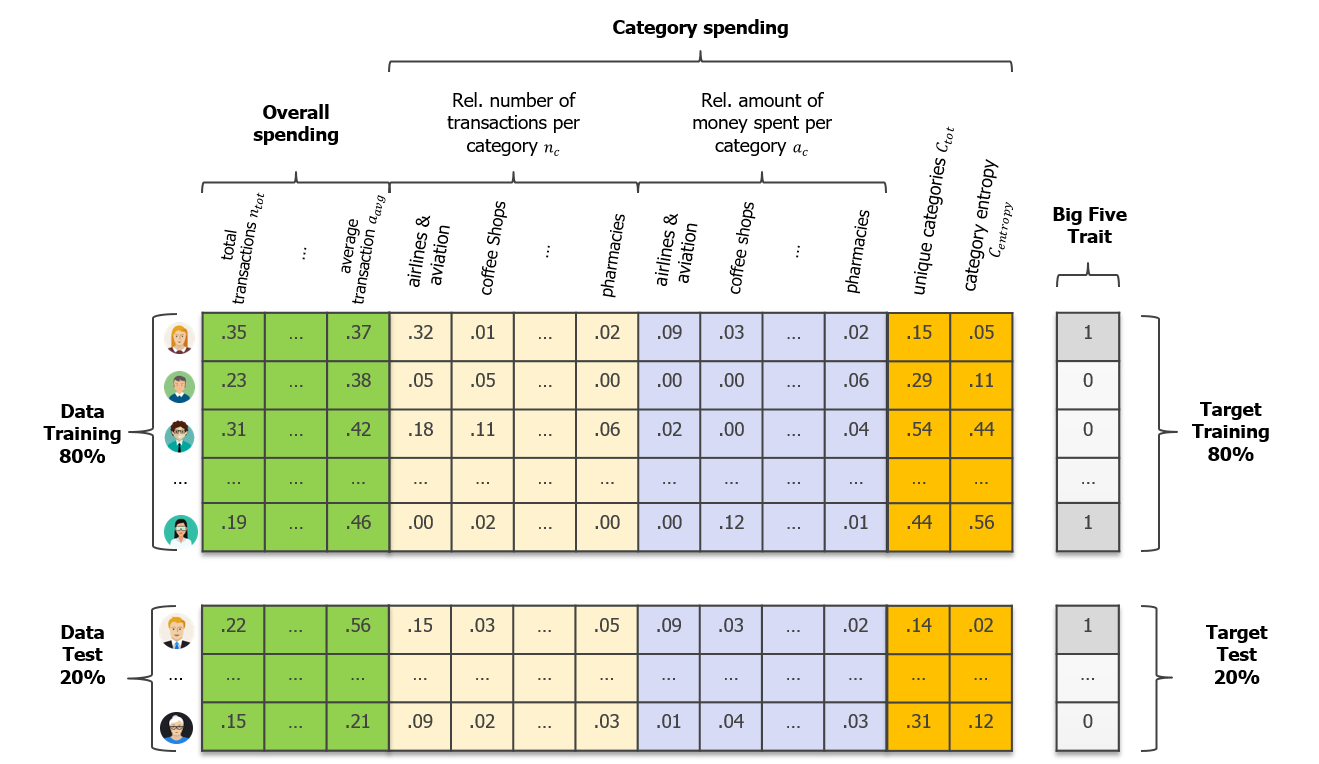}}
	\caption{Pre-processed financial transactions data and binary target variable (Big Five trait).}
	\label{fig:DataTarget}
\end{figure}

We train all models using the same financial transactions data to predict High and Low levels on the traits. \textbf{Fig. \ref{fig:DataTarget}} depicts the pre-processed financial transactions data and binary target.
For each trait, we construct two separate models, in line with a one-vs.-rest approach for multi-class problems: (1) a model that decides if someone scores High on a trait and (2) one that decides if someone scores Low on a trait. For example, for Extraversion, we train a model that predicts High Extraversion and another that predicts Low Extraversion.

\subsubsection{Evaluation \& Selection} 
\begin{figure}[h!]
	\centering
	\scalebox{0.6}{\includegraphics{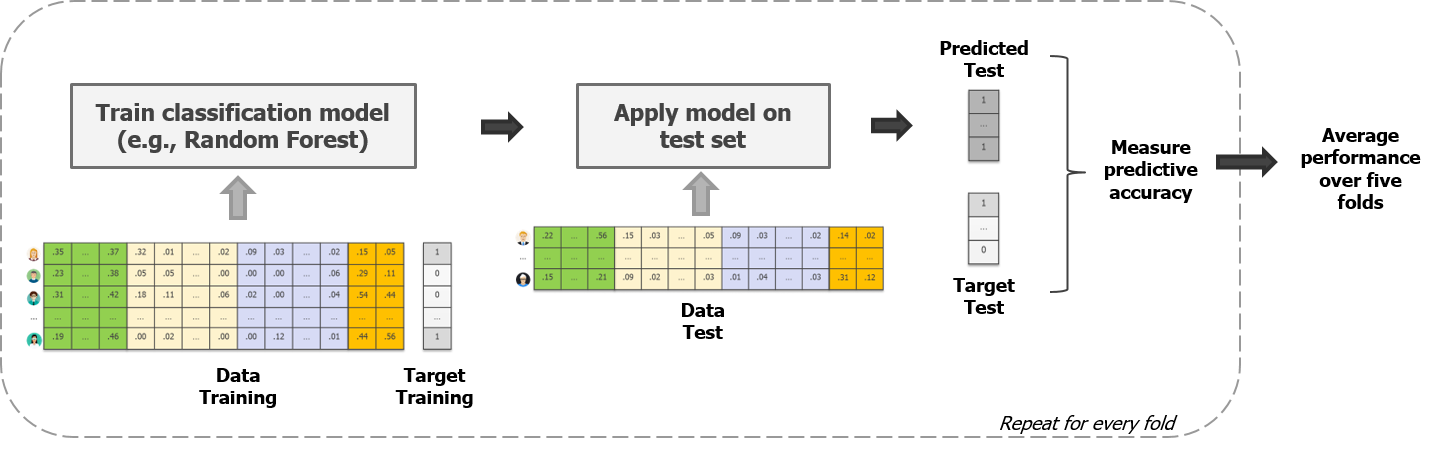}}
	\caption{Five fold cross-validation procedure to develop classification models to predict BF traits.}
	\label{fig:Modeling}
\end{figure}

The final classification system consists of a (continuous) scoring function $f$ that can be used to assign a score $s$ to every instance $\textbf{x}$. 
We use the \textit{Area under the Receiver Operating Curve (AUC)} to measure the general performance of the models. It reflects the model's ability to rank a true positive instance (e.g., a true extrovert) higher than a true negative instance (e.g., a true introvert)~\cite{DataScienceBusiness}. The AUC value does not depend on a classification threshold, but only on the score ranking of the instances that the model returns~\cite{DataScienceBusiness}. Moreover, AUC is not influenced by the underlying distributions of the personality classes (i.e., imbalance of the target variable). AUC is useful to summarize the model's performance in one metric and decouples classifier performance from the specific conditions under which the classifier will be used. Also, AUC allows for an easy comparison with random predictions, since a random classifier should result in a AUC value of $50\%$. We use AUC to compare the predictive accuracy of linear vs. nonlinear models and the predictability of different traits from the financial transactions data \cite{Phan2021}. We report the average AUC across the five folds (see \textbf{Fig. \ref{fig:Modeling}}). 

%Given a threshold $k$, explicit class predictions can be assigned to instances $\textbf{x}$ using a binary indicator function $\hat{\textbf{Y}}$$=$$I(s>k)$, with $f($\textbf{x}$)$$=$$s$, which, in turn, can be linked to a decision. After selecting the final model for each trait (see \textit{Linear vs. Nonlinear Models} in Section $3.1$), we obtain class label predictions using the final model and setting the threshold $k$ so that the fraction of individuals predicted as positive (i.e., belonging to a personality class) equals the fraction of positives in the data (approximately one-third of the data).
%We include four evaluation metrics to measure the model's performance at this threshold: $Accuracy$, $Precision$, $Recall$ and $Fscore$. The classifier $Accuracy$ reflects the proportion of correct classifications of the model.\footnote{$Accuracy$ has some well-known limitations, for example, it is misleading for imbalanced problems.
%Alternatively, we can look at both $Precision$ and $Recall$, or at the combined $Fscore$ measure, which are more suitable for imbalanced problems.} The \textit{Precision} measures the proportion of correct decisions among the predicted positive instances. $Recall$ measures the proportion of positives that are retrieved. The $Fscore$ is an evaluation metric that reflects the (harmonic) mean of $Precision$ and $Recall$, and is more suitable than $Accuracy$ for classification tasks with imbalanced target distributions.

\subsection{Model interpretability}

\subsubsection{Global explanations: CART to extract rules}
We use the \textit{CART} decision tree algorithm in \textit{Python} to extract global explanation rules. The algorithm extracts a set of if-then-else rules using the behavioral features together with the predicted classes $\hat{\textbf{Y}}$ of the classification model. We set the maximum tree depth to $3$, to limit the complexity of the explanation rules and make them easily understandable by humans. Depending on the setting, however, a user can increase the maximum complexity and get more granular explanations, with possibly additional insights.

\subsubsection{Local explanations: SEDC to compute counterfactual explanations}
We use the SEDC algorithm to compute (local) counterfactual rules (\textit{Python} code available\footnote{https://github.com/yramon/edc}), that is based on a best-first heuristic search strategy \cite{MartensProvostEDC,RamonCounterfactuals,Wachter}. We define counterfactuals as the set of features that need to change so that the predicted class changes, where a `change' is defined as replacing the original feature value with the median value of that feature computed over the training data.
To use SEDC, the decision-making (i.e., assignment of a person to a personality bucket) should be based on comparing a predicted score (i.e., the model's output) to a threshold. The scoring function is used by the SEDC algorithm so that it first considers features that, when replacing their value with the median, reduce the predicted score the most in the direction of the opposite class (i.e., the `best-first' feature).

\subsection{Results}
\label{Results}
In the following sections we will outline how XAI methods can be used to validate predictive models that compute personality from real-world transactions data. We first discuss the extent to which personality traits and facets can be predicted using both linear and non-linear models ($3.5.1$). Next, we show how rule extraction explains classifiers at an aggregate level and describe practical use cases of global interpretability on the basis of concrete examples from the case study ($3.5.2$ - Global explanations). Lastly, we provide empirical support for why local explanations are important---especially when modeling behavior---and elaborate on the implications of our observations ($3.5.2$ - Local explanations).

\subsubsection{Classification Performance Analysis}

\textit{Linear vs. nonlinear techniques.}

\begin{figure}[ht]
	\centering
	\scalebox{0.4}{\includegraphics{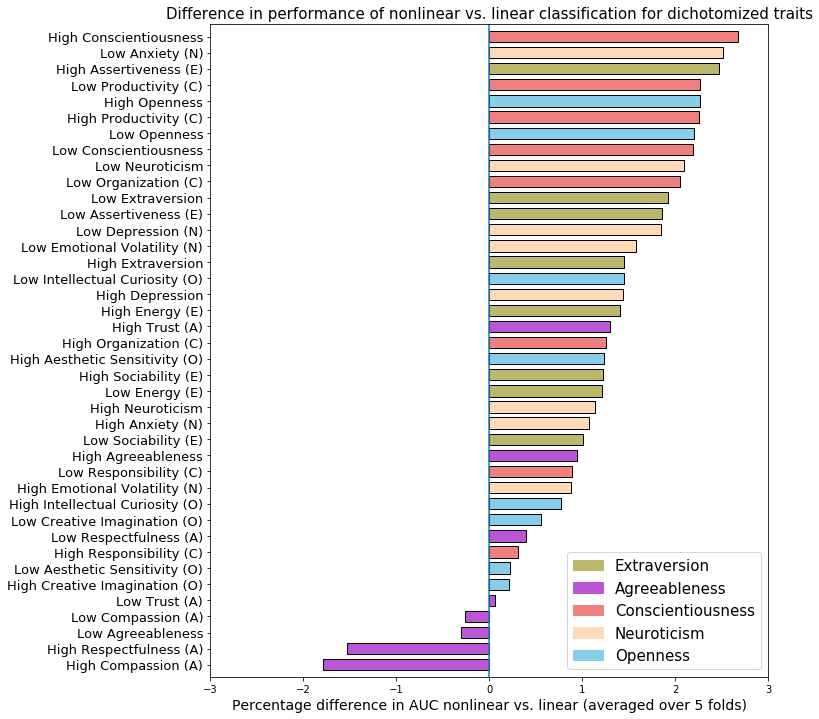}}
	\caption{Percentage difference in predictive accuracy of nonlinear vs. linear classification models for dichotomized personality traits, expressed by the difference in Area under the Curve (AUC), and ranked by decreasing difference in AUC. %The bars represent the percentage difference in AUC of nonlinear vs. linear models averaged over five folds of the cross-validation. 
	Positive values indicate that the best nonlinear model outperformed the best linear model.}
	\label{fig:AUC_LvsNL}
\end{figure}

First, we focus on the performance of linear vs. nonlinear techniques to model personality. We compare the performance of linear models (LR and Logit) vs. nonlinear models (RF), measured by the difference in AUC. The goal here is to provide a sound statement regarding the superiority of more flexible techniques for modeling personality from spending data. For the majority of traits and facets, nonlinear models outperform linear models (see \textbf{Fig. \ref{fig:AUC_LvsNL}}). On average, traits could be predicted with $58.14$\% accuracy in the linear models (min=$53.13$\%, max=$61.82$\%), and $59.31$\% in the nonlinear models (min=$53.35$\%, max=$63.98$\%).
Because we find that RF models---capable of finding nonlinear patterns--- generally outperform the linear models, we select RF as the final technique and report all following results based on the outputs of the RF models.

\hfill

\newpage
\textit{Predictability of personality traits and underlying facets.}

\begin{figure}[h!]
	\centering
	\scalebox{0.45}{\includegraphics{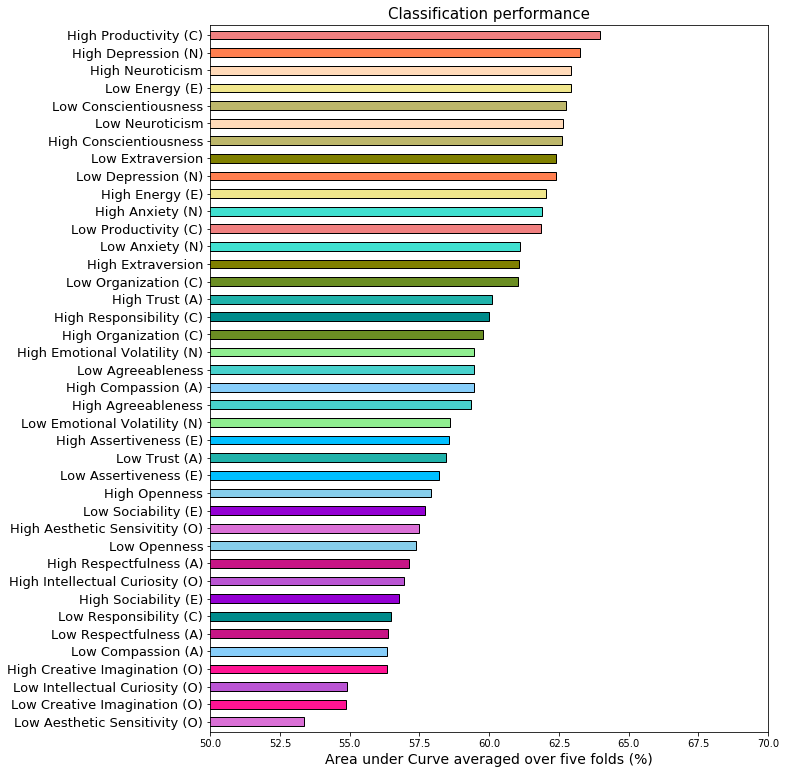}}
	\caption{\textit{Classification performance.} Prediction accuracy of models that classify High or Low levels of personality traits and facets expressed by the AUC averaged over five folds. The personality trait to which each facet belongs, is shown in parentheses (e.g., `E' stands for Extraversion).}
	\label{fig:AUC_5CV}
\end{figure}

\textbf{Fig. \ref{fig:AUC_5CV}} shows the prediction accuracy of the selected models that classify personality. There is a wide variation in the models' performances, ranging from moderate (e.g., AUC=$53.4$\% for Low Aesthetic Sensitivity) to decent performance (e.g., AUC=$63.9$\% for High Productivity). The best classification performance is achieved when predicting High levels of Productiveness, Depression and Neuroticism. Overall, individuals can be classified substantially above chance level for the majority of traits, which is in line with prior work that explored the value of spending data to segment people based on their personalities. The performances we find are comparable with, and even slightly better than, accuracies reported in related studies that use machine learning to predict BF traits from spending data \cite{TovanichSpending,SpendingPersonality}.

A second observation is that (the facets in) Conscientiousness and Neuroticism are the most predictable traits from the data, while Agreeableness and Openness characteristics are the least predictable. One possible explanation for this observation is that implicit behavioral residues---like the transaction records in this study---are particularly useful to predict \textit{intrapersonal} characteristics (Conscientiousness and Neuroticism), while other types of digital footprints that constitute more explicit identity claims, like social media data, are more valuable for recognizing \textit{interpersonal} traits \cite{Pianesi2008,Blom2011} (Openness, Extraversion and Agreeableness). Our results suggest that the spending patterns differ more between those groups scoring different on intrapersonal traits, allowing for a better classification compared to interpersonal traits.

Further, the facets that underlie the same trait are not always equally predictable. For example, it is easier to predict Energy levels from financial transactions than Sociability and Assertiveness (all facets of Extraversion). Similarly, high levels of Productivity are easier to predict than Low levels, and Emotional Volatility as part of Neuroticism is less predictable from these data than Anxiety and Depression.

% Figure

\subsubsection{Model Interpretability Analysis}
In the next sections we explore the explainability of the models at the global and local level and discuss use cases of interpretability on the basis of examples from our case study. For simplicity, we only explain predictions of the Random Forest models that predict High levels of a trait which are more frequently used in applied contexts. However, the analysis would follow a similar pattern for the models that predict Low levels. Our goal is to demonstrate the value and different use cases of XAI by means of a realistic case study. We aim to provide compelling evidence to academics and practitioners for the importance of XAI methods in \textit{any} application that leverages behavioral data to assess psychological traits, making the implications of our findings relevant beyond the examples presented in this case study.

\hfill

\textit{Global explanations: rule extraction.}

\begin{figure}[h!]
	\centering
	\scalebox{0.5}{\includegraphics{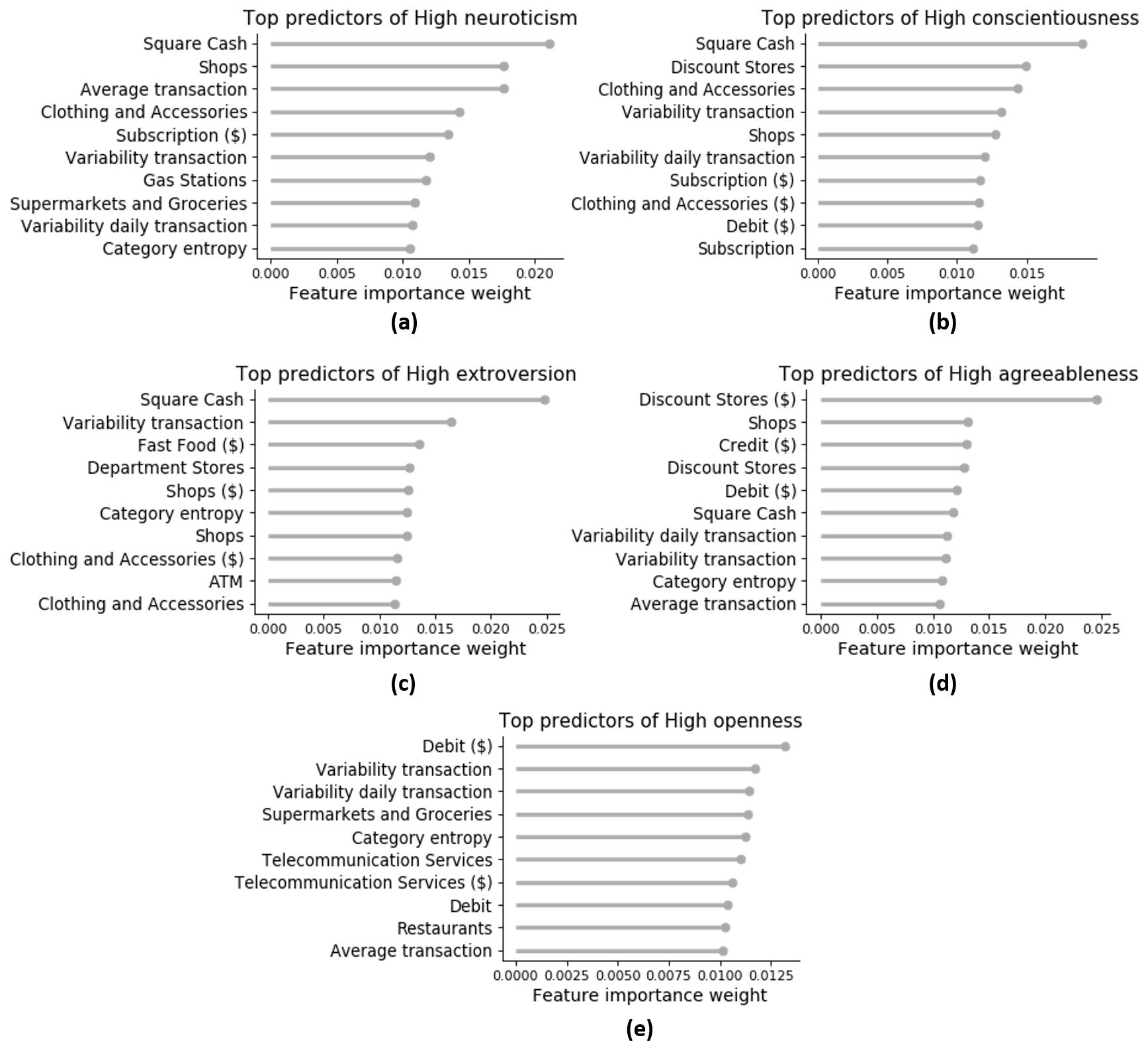}}\label{fig:a}
	\caption{\textit{Feature relevance lists.} Top features in the model that predicts High levels of (a) Neuroticism, (b) Conscientiousness, (c) Extraversion, (d) Agreeableness, and (e) Openness. The importance weights are computed as the average impurity reduction over the trees in the Random Forest.}
	\label{fig:feature_relevance}
\end{figure}

\textbf{Tables \ref{table:rules}} and \textbf{\ref{table:performance_rules}} respectively show the explanation rules that approximate the classification behavior of the models that predict personality and their quality. The predictions of the rules substantially overlap with the model predictions ($Fidelity$ ranges from $72.07$\% to $81.59$\%) and the rules that explain when a trait is predicted achieve high levels of reliability (see the $Precision_{f}$ column in \textbf{Table \ref{table:performance_rules}}). When comparing the rules and the feature relevance lists (shown in \textbf{Fig. \ref{fig:feature_relevance}}), we observe a considerable amount of overlap of the top features identified as important in the black box model. However, the feature relevance lists do not explain how the feature values lead to a classification of interest, and cannot account for interactions of features or shed light onto the directionality of the effects. In contrast, the extracted rules displayed in \textbf{Table \ref{table:rules}} capture associations between features and personality classes that the model learned and utilized in the prediction task.

To make the decision rules more tangible, we discuss a number of face-valid examples that are representative of these global explanations (see \textbf{Table \ref{table:rules}}).  Focusing on the personality trait of Conscientiousness, for example, the explanation shows that individuals with high transaction volumes in Discount stores %and low variability in transaction amount 
are more likely to be classified as conscientious by the algorithm. This rule %nicely 
aligns with the general description of conscientiousness as the tendency to exercise self-control and to be less impulsive. In addition, the model identified the association between Conscientiousness and high transaction volumes in Clothing \& Accessories and Beauty products, which is consistent with research showing that conscientious individuals demonstrate a stronger interest in clothing and physical appearance than individuals scoring low on Conscientiousness. \cite{Aiken1963,Darden1975,TovanichSpending}. Moreover, the rules provide insight into specific model behavior, such as trade-offs made by the model to make personality classifications which cannot be identified in the feature relevance list (see \textbf{Fig. \ref{fig:feature_relevance}}). More precisely, the rules show that the model classifies someone as conscientious when there are many transactions in the categories Square Cash and Beauty products, irrespective of spending volumes in other categories. However, when a person's relative spending in the Beauty products category drops below a certain threshold ($0.3$\%), then a substantial amount of spending in the category Clothing \& Accessories needs to be observed to still classify the person as conscientious.

Gaining insight into how predictors impact personality classifications at a global level can also help explore new hypotheses about the relationship between spending behavior and psychological traits. In our case study, it is notable that, within the money transactions space, there are different payment services that are predictive for different personalities. This can trigger new research questions, such as, why a specific group of people---homogeneous in terms of personality--- would develop their own distinct taste in payment services (e.g., see research on brand personality). More precisely, an important category in the models to predict personality is Square Cash, a mobile payment application that allows users to easily transfer money to friends and family. Since this mobile application is identified as important for explaining classifications of the algorithm (Square Cash appears in almost all explanations in \textbf{Table \ref{table:rules}}), future research might investigate this relationship to understand what makes Square Cash users uniquely conscientious or not depending on its interaction with other spending features.

% Table
\begin{table}[h!]
	%\centering
	\caption{\textit{Global explanation rules.} If-then-else rules that explain when the algorithm classifies High levels of personality traits based on financial transactions. The Default class comprises Low to Medium levels of the same trait. Note: Discount stores and Discount stores (\$) respectively indicate the relative number of transactions in vs. the amount of money spent in a category. `Square Cash' and `Venmo' are mobile payment applications to transfer money to friends and family.
	}
	\label{table:rules}
	\scalebox{0.85}{
		\begin{tabular}{c|l}
		    \\
			\textbf{Trait} & \textbf{Explanation rules}\\
			\noalign{\smallskip}\hline
		%	\parbox[t]{2mm}{\multirow{5}{*}{\rotatebox[origin=c]{90}{Neurotic}}}\\
	    	 & \textit{if (Square cash(\$) $\leq$ 0.3\%) and (Average transaction $\leq$ \$57.08) and (Clothing \& Accessories $\leq$ 0.7\%)  \textrightarrow \ Model predicts High Neuroticism}\\
			N & \textit{if (Square cash(\$) $>$ 0.3\%) and (Subscription(\$) $>$ 0.5\%) and (Loans \& Mortgages(\$) $\leq$ 3.9\%)  \textrightarrow  \ Model predicts High Neuroticism}\\
		    & \textit{else: Model predicts Default}\\
		    \\
		    \noalign{\smallskip}\hline
		    %\parbox[t]{2mm}{\multirow{6}{*}{\rotatebox[origin=c]{90}{Conscientious}}}\\
			& \textit{if (Square cash $>$ 0.4\%) and (Beauty Products $>$ 0.3\%)  \textrightarrow \ Model predicts High Conscientiousness}\\
			C & \textit{if (Square cash $>$ 0.4\%) and (Beauty Products $\leq$ 0.3\%) and (Clothing \& Accessories(\$) $>$ 0.8\%)  \textrightarrow  \ Model predicts High Conscientiousness}\\
			& \textit{if (Square cash $\leq$ 0.4\%) and (Discount Stores $>$ 0.8\%) and (Shops $>$ 0.5\%) \textrightarrow \ Model predicts High Conscientiousness}\\
		    & \textit{else: Model predicts Default}\\
		    \\
			\noalign{\smallskip}\hline
			%\parbox[t]{2mm}{\multirow{5}{*}{\rotatebox[origin=c]{90}{Extroverted}}}\\
		    & \textit{if (Square cash $\leq$ 0.7\%) and (Clothing \& Accessories (\$) $>$ 0.7\%) and (Hotels \& Motels $>$ 0.1\%)  \textrightarrow \ Model predicts High Extraversion}\\
		    E & \textit{if (Square cash $>$ 0.7\%) and (Variability transaction amount $\leq$ 0.31) \textrightarrow \ Model predicts High Extraversion}\\
		    & \textit{if (Square cash $>$ 0.7\%) and (Variability transaction amount $>$ 0.31) and (Service $>$ 0.3\%) \textrightarrow \ Model predicts High Extraversion}\\
		    & \textit{else: Model predicts Default}\\
		    \\
		    \noalign{\smallskip}\hline
		    %\parbox[t]{2mm}{\multirow{5}{*}{\rotatebox[origin=c]{90}{Agreeable}}}\\
            & \textit{if (Square cash $\leq$ 0.5\%) and (Discount Stores(\$) $>$ 0.1\%) and (Shops $\leq$ 0.6\%)  \textrightarrow \ Model predicts High Agreeableness}\\ 
            A & \textit{if (Square cash $>$ 0.5\%) and (Discount Stores $>$ 0.7\%) \textrightarrow \ Model predicts High Agreeableness}\\
            & \textit{if (Square cash $>$ 0.5\%) and (Discount Stores $\leq$ 0.7\%) and (ATM $>$ 5.7\%) \textrightarrow \ Model predicts High Agreeableness}\\
		    & \textit{else: Model predicts Default}\\
		    \\
		    \noalign{\smallskip}\hline
		    %\parbox[t]{2mm}{\multirow{5}{*}{\rotatebox[origin=c]{90}{Open}}}\\
            & \textit{if (Venmo(\$) $>$ 0.1\%) \textrightarrow \ Model predicts High Openness}\\ 
            O & \textit{if (Venmo(\$) $\leq$ 0.1\%) and (Square cash(\$) $>$ 0.5\%) and (Digital purchase $>$ 2.5\%) \textrightarrow \ Model predicts High Openness}\\ 
            & \textit{if (Venmo(\$) $\leq$ 0.1\%) and (Square cash(\$) $\leq$ 0.5\%) and (Taxi(\$) $>$ 0.4\%) \textrightarrow \ Model predicts High Openness}\\ 
		    & \textit{else: Model predicts Default}\\
	    \end{tabular}}
\end{table}

\begin{table}[h!]
	\centering
	\caption{\textit{Quality of explanation rules}. Out-of-sample performance of rules that explain the model's classifications. The performance of a random explanation is shown in parentheses.
	}
	\label{table:performance_rules}
	%\scalebox{0.85}{
		\begin{tabular}{ccccc}
		    \\
			\textbf{Personality class} & \textbf{$Fidelity$ (\%)} & \textbf{$Fscore_{f}$ (\%)} &  \textbf{$Precision_{f}$ (\%)} & \textbf{$Recall_{f}$ (\%)}\\
			\noalign{\smallskip}\hline
			Neuroticism & $79.02$ ($58.16$) & $62.48$ ($29.79$) & $66.87$ ($29.79$) & $58.64$ ($29.79$)\\
			Conscientiousness & $75.82$ ($58.74$) & $52.45$ ($29.09$) & $61.29$ ($29.09$) & $45.84$ ($29.09$)\\
			Extraversion & $78.47$ ($55.57$) & $58.43$ ($33.31$) & $81.86$ ($33.31$) & $45.43$ ($33.31$)\\
			Agreeableness & $81.59$ ($60.95$) & $63.35$ ($26.59$) & $67.33$ ($26.59$) & $59.82$ ($26.59$)\\
			Openness & $72.07$ ($56.78$) & $50.82$ ($31.59$) & $57.28$ ($31.59$) & $45.68$ ($31.59$)\\
	\end{tabular}
\end{table}

% Figure

Lastly, global model interpretability can help identify problems or weaknesses of the model, for example, related to the data quality or the generalizability of the model. When modeling human behavior, monitoring the performance of a model and understanding the contribution of individual (behavioral) features can be crucial. For example, changes in the meaning of certain behaviors can result in sudden drops in performance over time, a phenomenon termed `concept drift' (described in Section \ref{Introduction}). Returning to the the mobile application Square Cash, for example, it is conceivable that such a mobile applications might at first be niche product that is only used by specific groups with similar psychological profiles, but over time becomes more widespread and used by a wider population. As a result, the spending feature might lose its predictive power, challenging the expected lifetime of the prediction model.

\hfill

\newpage
\textit{Local explanations: counterfactual explanations.}

In addition to global model interpretability, we compute local explanations to identify important features for individual classifications. In \textbf{Table \ref{table:CF_Neuroticism_Median}}, local explanations are shown for why individuals who are predicted to be highly neurotic. For example, the explanation for why Person $E$ was predicted to be neurotic can be interpreted as follows: ``if Person E had spent \textit{less money} in Department Stores, but \textit{more frequently} in Square Cash $\rightarrow$ then Person E would not have been predicted to be neurotic''. 
There are some interesting observations when looking at the counterfactuals in \textbf{Table \ref{table:CF_Neuroticism_Median}}. 
First, our experiments show that the explanations are generally concise (on average, explanations consist of $0.3$\% of the full feature space). 

\begin{figure}[ht]
	\centering
	\scalebox{0.5}{\includegraphics{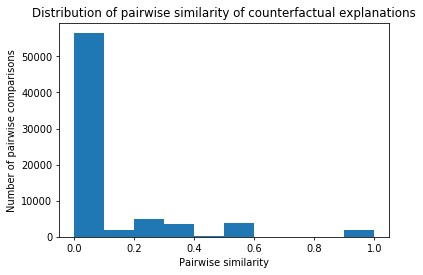}}
	\caption{\textit{Similarity of explanations.} Distribution of pairwise similarity between counterfactual explanations for predicting Neuroticism. A value of $0$ (resp., $1$) indicates no (resp., perfect) overlap.}
	\label{fig:pairwise}
\end{figure}

Second, the explanations vary tremendously in nature: People are assigned to the same personality class based on vastly different behaviors. In other words, there is a lot of uniqueness in the explanations associated with each individual. This is visually depicted by \textbf{Fig. \ref{fig:pairwise}} which plots the distribution of pairwise similarities between counterfactual explanations. We observe that the majority of explanations has no overlap. This observation is consistent with prior work on local explanations for models on behavioral data demonstrating the variety of local explanations \cite{MartensProvostEDC,ChenCloaking,RamonCounterfactuals}.

When explaining the predictions for the personality trait of Neuroticism, $91.1$\% of the explanations are unique. This implies that people will receive different explanations most of the time. As a result, the local explanations provide insights into the specific behavior of a person that led the model to make a decision, making the explanation more granular and personally relevant than the global explanation rules. To illustrate this more clearly, consider two female individuals in our sample, both classified as neurotic by the model. Examining the global explanations in \textbf{Table \ref{table:rules}}, they are both explained by the first explanation rule, that includes the features Square Cash, Average transaction amount and Clothing \& Accessories.\footnote{Note that the global explanation shows which combination of feature values likely leads the model to predict a Neurotic person, however, it does not give an exhaustive ($Recall_{f}$ is not $100$\%) nor perfectly reliable ($Precision_{f}$ is not $100$\%) rule set that explains when the model predicts a Neurotic person. Moreover, changing the features' values such that the rule would no longer apply to the person, does not guarantee that the predicted class flips to the Default, because there might be other combinations of feature values---not captured by the incomplete global explanation---that lead to the prediction of a Neurotic person.} However, going a level deeper to the local explanations, we get a more granular notion of which features contributed to the classification of each of the two women. For the first woman, the predictors Gas Stations, Square Cash and Taxi are part of the explanation for being classified as neurotic. In contrast, the second woman would receive an explanation that comprises the features Average transaction amount, Clothing \& Accessories, Fast Food and Public Transportation Services.

\begin{figure}[ht]
	\centering
	\scalebox{0.3}{\includegraphics{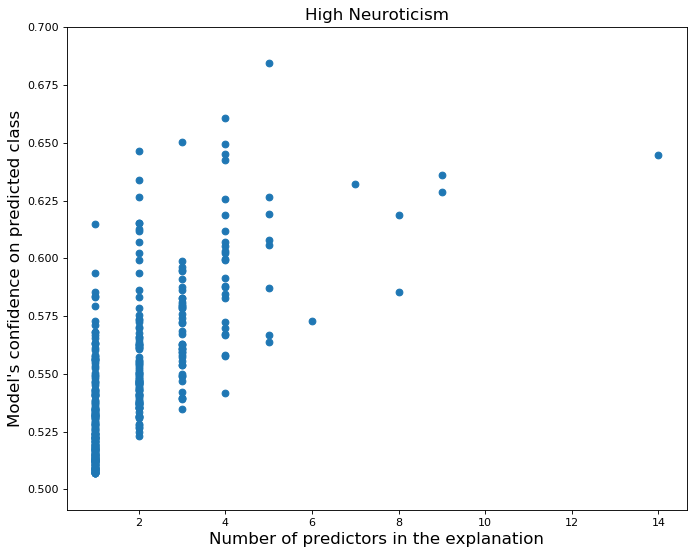}}
	\caption{\textit{Model's classification confidence vs. size of counterfactual explanations (High Neuroticism).} Model's predicted score vs. number of predictors in the explanation to counterfactually explain the predicted class. The correlation between the scores and explanation sizes is $.68$.}
	\label{fig:ConfvsSize}
\end{figure}

Third, local explanations not only vary in the specific features and feature combinations they use, but also in the complexity of the counterfactual rules to explain decisions. Depending on someone's set of historical transactions (their `financial behavior profile'), it can become harder to flip the model's predicted class. Generally, in the results, we observe a trend that the number of features that counterfactually explain the predicted class positively relates to the prediction confidence of a model as depicted by \textbf{Fig. \ref{fig:ConfvsSize}}. Moreover, the number of feature changes needed to flip the predicted class is generally larger for True Positives compared to False Positives. This finding provides some intuitive satisfaction and is in line with prior work on counterfactual explanations \cite{ChenCloaking}. When explaining why individuals are predicted as neurotic, the average number of features in the explanations for True Positives and False Positives is respectively $2.09$ and $1.79$. This difference suggests that a person who is incorrectly classified as neurotic needs to change fewer features to receive a different classification than someone who was accurately classified to be neurotic.

Finally, the explanations in \textbf{Table \ref{table:CF_Neuroticism_Median}} provide another interesting insight. For example, Person $A$ was predicted as neurotic due to two features: ``if Person A had spent \textit{more frequently} in Clothing \& Accessories and Restaurants, but \textit{less frequently} in Computers \& Electronics, Insurance and Shops $\rightarrow$ then Person A would not have been predicted as neurotic''. The rule highlights that it is not always the behavior that people exhibit that are most predictive for a psychological characteristic. The behavior that people do not or only rarely exhibit might also drive the model's classification.

% Figure

% Figure

% Table
\begin{table}[h!]
	\centering
	\caption{\textit{Counterfactual explanations for predicting Neuroticism.} Local explanations that show the features that (counterfactually) explain the predicted class High Neuroticism. A selection of explanations is shown for instances $i$ with highest predicted scores $s_{i}$.
}
	\label{table:CF_Neuroticism_Median}
	\scalebox{0.85}{
		\begin{tabular}{ll}
		    \\
		    \textbf{Instance $i$} & \textbf{Counterfactual explanation for instance $i$}\\
		    \hline
		    Person a ($s_{a}$=.69) & If you had spent \textit{less frequently} in Computers \& Electronics, Insurance and Shops,\\
		    $size_{CF,a}$$=$$5$ & and \textit{more frequently} in Clothing \& Accessories and Restaurants \textrightarrow then you would not\\
		    & have been predicted as Neurotic\\
		    \hline
		    Person b ($s_{b}$=.66) & If you had spent \textit{less frequently} in Pets, Shops and Veterinarians, and spent \textit{less}\\
		    $size_{CF,b}$$=$$4$ & \textit{money} on Subscription \textrightarrow you would not have been predicted as Neurotic\\
		    \hline
		    Person c ($s_{c}$=.65) & If you had spent \textit{less frequently} in Shops, \textit{less money} on Internal Account Transfer\\
		    $size_{CF,c}$$=$$3$ & and Subscription \textrightarrow then you would not have been predicted as Neurotic\\
		    \hline
		    Person d ($s_{d}$=.65) & If you had spent \textit{less frequently} in Shops, and \textit{less money} on Subscription\\
		    $size_{CF,d}$$=$$2$ & \textrightarrow then you would not have been predicted as Neurotic\\
		    \hline
		    Person e ($s_{e}$=.65) & If you had spent \textit{less frequently} in Food \& Beverage, PayPal and Shops, and \textit{less}\\
		    $size_{CF,e}$$=$$4$ & \textit{money} on Subscription \textrightarrow then you would not have been predicted as Neurotic\\
		    \hline
		    Person f ($s_{f}$=.65) & If you had spent \textit{less frequently} in Check, Department stores and Shops, and \textit{more}\\
		    $size_{CF,f}$$=$$4$ & \textit{frequently} in Supermarkets \& Groceries \textrightarrow then you would not have been predicted\\
		    & as Neurotic\\
		    \hline
		    Person g ($s_{g}$=.64) & If you had spent \textit{less frequently} in Shops and Tobacco, and \textit{less money} on\\
		    $size_{CF,g}$$=$$4$ & Subscription and Tobacco \textrightarrow then you would not have been predicted as Neurotic\\
		    \hline
		    Person h ($s_{h}$=.64) & If you had spent \textit{less frequently} in Food \& Beverage, Vintage \& Thrift, \textit{less money} on\\
		    $size_{CF,h}$$=$$8$ & Department stores, Shops, Tobacco and Vintage \& Thrift, \textit{more frequently} in Clothing \&\\
		    & Accessories, \textit{more money} in Arts \& Entertainment, and the variability of your spending\\
		    & amount was \textit{lower} \textrightarrow then you would not have been predicted as Neurotic\\
	\end{tabular}}
\end{table}

\newpage
\section{Discussion}
\label{Discussion}

In this paper, we demonstrated the value of XAI in the context of psychological profiling that translates innocuous digital footprints into psychological traits. Our case study highlights the importance of both global and local methods to address interpretability challenges when working with high-dimensional, sparse, behavioral data.

\subsection{Importance of global explanations and implications}
Global rules provide general insights into the decisions a model makes about a target based on what it has learned from the the full (training) data set. Global rules hence provide an explanation of the decision model that is comprehensible to the individuals making predictions and the individuals who are the target of predictions \cite{MartensEthics}. While other global XAI methods exist (e.g. feature relevance scores), we argue that rule extraction---and surrogate explanations in general---is a particularly useful tool to understand how a (combination of) feature(s) impact(s) model \textit{decisions}, and to provide an estimate of how well the classifications can be explained (measured by $Fidelity$).
Insight into the $Fidelity$ of an explanation is important in the context of behavioral data. If the most important features in a model are extremely sparse, an explanation with few rules and/or few conditions per rule will fail to make accurate predictions for most people, as reflected by a low $Fidelity$ or $Fscore_{f}$. When this is the case, novel rule extraction approaches can be used to replace features with metafeatures (groups of individual behavioral features - e.g., "fast food" purchases that are made up of individual merchants) to increase the $Fidelity$ of the extracted rules \cite{RamonMetafeatures}. 

Our experiments demonstrate how global rules can be used to validate what the model learned at an aggregate level. This additional understanding can add a layer of trust to the out-of-sample performance measures by testing the face validity of the global rules (i.e., compare them with related work and existing knowledge). Not only could global rules be used to validate models before they are deployed in practice, but they could also be used to continuously audit the functionality of a model (i.e., does it use information that we do not \textit{want} it to use?). For example, when verifying if a model exhibits algorithmic bias toward a certain protected group (e.g., minorities), we can use post hoc explanations to audit the model. Importantly, global rule extraction calls for the inclusion of domain experts. In our case study, for example, personality psychologists can help determine whether the extracted rules make sense in the context of the vast body of literature on the correlates of personality traits.

Our case study also shows how XAI at a global level can be used to generate novel hypotheses that would have been impossible to derive deductively (e.g., different preferences for mobile payment services). As \cite{Stachl2020} note, researchers should ``invest time and effort to finding persistent and stable digital behavioural dimensions when working on theoretical models''. A future direction that is worth exploring is how higher-level, less-sparse metafeatures can help construct more `stable' behavioral profiles that can be used for (bottom-up) theory building and hypothesis generation. This is especially interesting when modeling very high-dimensional and sparse behavioral data, for example, modeling the fine-grained places people visit \cite{MullerPeters}, web pages they browse, or pages they `like' on Facebook \cite{Kosinski2013}. While individual places, websites and Facebook pages might be highly predictive at any given point in time, they are also likely to change (e.g. the same coffee shop, website or Facebook page might only survive for a certain period of time). Using metafeatures for modeling is likely to lead to worse predictive performance in the moment \cite{DeCnuddeBenchmark, JunqueBiggerBetter, ClarkProvost}. However, they might prove valuable when extracting insights from high-performing models and generating hypotheses that are more stable over time.
In sum, global rule extraction methods provide researchers and practitioners with a tool to validate their models, create a more robust foundation for future investigations of the relationship between human behavior and psychological constructs, and facilitate replication efforts in computational social sciences research~\cite{OpenScience}.

\subsection{Importance of local explanations and implications}
Next to global insights, our experiments highlight the importance of local counterfactual rules to address interpretability issues of models on behavioral data. While global rule extractions have partially found their way into social science research, local counterfactuals (and other types of local explanations) have largely been overlooked so far. The value of local rule extractions is manifold. First, they are concise: Only a small fraction of features of the full feature space is part of the explanations. We might worry providing users with explanations that are unnecessarily large, especially in the context of behavioral data. In our experiments, we see that the explanations generally have a small size, especially relative to the total number of features present in the model. This concurs with findings of \cite{ChenCloaking}, \cite{Fernandez} and \cite{MartensProvostEDC}. 

Second, they are specific to the individual's behavior: Explanations point at \textit{unique} behavior of the person that contributed most to the classification. Counterfactual explanations have the additional advantage that they are consistent with requirements currently described in regulation \cite{Wachter}. For example, the advisory organ of the EU on GDPR, Working Party 29, provided additional details on meaningful information that data subjects should receive when subject to automated decisions: ``The company should find simple ways to tell the data subject about the rationale behind, or the criteria relied on in reaching the \textit{decision},...The information should be sufficiently comprehensive for the data subjective to understand the reasons for the \textit{decision}.'' Local rule extractions satisfy these requirements.

Third, local explanations could be used by companies that chose to be transparent about the ways by which they target individuals. In addition to mandated regulations, Facebook's `Why Am I Seeing This Ad' initiative or the AdChoices program \cite{ChenCloaking}, for example, could provide their users with a clearer and more personalized explanation for why they are seeing a given ad. Notably, prior work has suggested that contrary to most people's intuition, transparency and control in the context of online advertising can indeed result in higher engagement levels \cite{Aaker1997,Tucker2014}. Similarly, local rule extraction might solve a problem many companies are facing when sharing global rules with users. Especially when the stakes are high, there is a concern that individuals will use these insights to `game the system'. In the context of hiring or lending for example, companies often do not want to disclose the exact working of the predictions they make, because they are worried that their models will be rendered inadequate as soon as individuals have the ability to strategically update their records in a certain direction. Local explanations are inherently relevant to one individual, but not per se useful for other people. This greatly reduces the risk of `gaming' the system.

Finally, for experts interacting with a model (e.g., psychologists or HR managers), local explanations can be useful to understand how a particular prediction was made, and what to focus their attention on: either to overrule the decision (when domain knowledge or context outweighs the explanation for the decision), or to understand a misclassification to guide error analysis. For example, consider the (fictitious) example of predicting mental health problems from online web searches. When a person is identified as depressed by the algorithm, it is useful for experts to validate the decision based on the (words in) searches that contributed most to this decision, instead of going through the hundreds of searches of this person. Further, knowing why the classification was made can be used to overrule the decision: say a person was identified as depressed because of searching for `symptoms of depression', but, when looking at queries in the same time window, it turns out the person logged in multiple times on the web page of the Department of Psychology at Columbia University. A user seeing this explanation would better understand why the prediction was made, and in this case, likely identify it as a false positive prediction, as this might not be an unusual search query for Psychology students.

\section{Conclusions}
\label{Conclusions}
Psychological profiling from digital footprints has attracted considerable interest from researchers and practitioners alike who study and apply the methodology across a wide variety of applications, ranging from marketing to employment to healthcare. Given that the underlying models can become very complex, they have earned the reputation of being a `black box' that is difficult to penetrate. Most of the research in this area has focused on the predictive accuracy of models, without much effort being dedicated to explaining how the classifications come about. However, the explainability of these systems---central in the field of Explainable AI---is becoming an essential requirement to generate trust and increase the acceptance of predictive technologies as well as generate better insights from these systems.

In this study, we showed how global and local XAI techniques can help domain experts and data subjects validate, question, and improve models that classify psychological traits from digital footprints. Using real-world financial transactions data to predict Big Five personality traits, we demonstrated how global rule extraction can be used to understand a model's classification behavior at an aggregate level, and discussed use cases of global model interpretability (validation, insights and improvement). Furthermore, we showed how local counterfactual rules can reveal more granular insights into why classifications are made (i.e., individuals are classified as exhibiting a personality trait for reasons that reflect their unique financial spending behavior), and discussed implications of this uniqueness for experts and data subjects. We hope this study encourages researchers and practitioners in the field of psychological profiling to implement XAI as a tool that can help them develop more human-centric, interpretable psychological profiling systems that support decision-making.

\section{Appendix}
\label{Appendix}

\begin{table}[h!]
	\centering
	\caption{Mean and standard deviation of the BF traits and facets in this study vs. the Internet sample \cite{BFI2S}. The fourth column shows the mean-level difference $d$ between the two samples. The last column represents the Cronbach's alpha of each item scale that measures a BF trait.}
	\label{table:Internet_sample}
	\scalebox{0.9}{
		\begin{tabular}{ccccc}
			\textbf{Domain or facet} & \textbf{This study} & \textbf{Internet sample} & $d$ & \textbf{Cronbach's alpha}\\
			\noalign{\smallskip}\hline
			Extraversion & $3.35$ ($1.08$) & $3.23$ ($.80$) & $.12$ & $0.8012$\\
			\textit{Sociability} & $3.21$ ($1.08$) & $2.95$ ($1.05$) & $.26$\\
			\textit{Assertiveness} & $3.58$ ($1.02$) & $3.28$ ($.93$) & $.30$\\
			\textit{Energy} & $3.25$ ($1.13$) & $3.47$ ($.89$) & $-.22$\\
			\noalign{\smallskip}\hline
			Agreeableness & $4.19$ ($.67$) & $3.68$ ($.64$) & $.51$ & $0.7868$\\
			\textit{Compassion} & $4.34$ ($.83$) & $3.84$ ($.78$) & $.50$\\
			\textit{Respectfulness} & $4.40$ ($.76$) & $3.98$ ($.71$) & $.42$\\
			\textit{Trust} & $3.84$ ($.92$) & $3.23$ ($.82$) & $.61$\\
			\noalign{\smallskip}\hline
			Conscientiousness & $3.87$ ($.79$) & $3.43$ ($.77$) & $.44$ & $0.8153$\\
			\textit{Organization} & $3.52$ ($1.17$) & $3.42$ ($1.01$) & $.10$\\
			\textit{Productivity} & $3.89$ ($.96$) & $3.37$ ($.90$) & $.52$\\
			\textit{Responsibility} & $4.18$ ($.82$) & $3.48$ ($.81$) & $.70$\\
			\noalign{\smallskip}\hline
			Neuroticism & $2.88$ ($.96$) & $3.07$ ($.87$) & $-.19$ & $0.8533$\\
			\textit{Anxiety} & $3.34$ ($1.07$) & $3.43$ ($.93$) & $-.09$\\
			\textit{Depression} & $2.61$ ($1.13$) & $2.85$ ($1.02$) & $-.24$\\
			\textit{Emotional volatility} & $2.67$ ($1.17$) & $2.93$ ($1.05$) & $-.26$\\
			\noalign{\smallskip}\hline
			Openness & $3.75$ ($.68$) & $3.92$ ($.65$) & $-.17$ & $0.7219$\\
			\textit{Intellectual curiosity} & $3.83$ ($.79$) & $4.10$ ($.70$) & $-.27$\\
			\textit{Aesthetic sensitivy} & $3.57$ ($.96$) & $3.80$ ($.92$) & $-.23$\\
			\textit{Creative imagination} & $3.83$ ($.94$) & $3.85$ ($.81$) & $-.02$\\
			\noalign{\smallskip}\hline
			& $N$$=$$6,408$ & $N$$=$$1,000$\\
	\end{tabular}}
\end{table}

\begin{table}[h!]
	\centering
	\caption{Summary of features capturing spending behavior.}
	\label{table:FeatureEngineering}
	\scalebox{0.7}{
		\begin{tabular}{cccc}
			\textbf{Type} & \textbf{Feature notation} & \textbf{Feature name} & \textbf{Description}\\
			\noalign{\smallskip}\hline
			Overall & $n_{tot}$ & Total transactions & Total number of transactions over $12$ months\\
			& $a_{tot}$ & Total amount transactions & Total amount of money spent over $12$ months\\
			& $a_{avg}$ & Average transaction & Average amount of money spent per transaction\\
			& $a_{cv}$ & Variability transaction & Variability of amount of money spent per transaction\\
			& $a_{avg,daily}$ & Average daily transaction & Average amount of money spent on a daily basis\\
			& $a_{cv,daily}$ & Variability daily transaction & Variability of amount of money spent on a daily basis\\
			\noalign{\smallskip}\hline
			Category & $n_{c}$ & Category $c$ & Relative number of transactions in category $c$ (e.g., Fast Food)\\
			& $a_{c}$ & Category $c$ (\$) & Relative amount of money spent in category $c$ (e.g., Fast Food (\$))\\
			& $C_{tot}$ & Unique categories & Number of distinct spending categories\\
			& $C_{entropy}$ & Category entropy & Diversity of spending in different categories\\
	\end{tabular}}
\end{table}

% *** Fig. ***

% *** Fig. ***

% *** Fig. ***
\begin{figure}[ht]
	\centering
	\scalebox{0.5}{\includegraphics{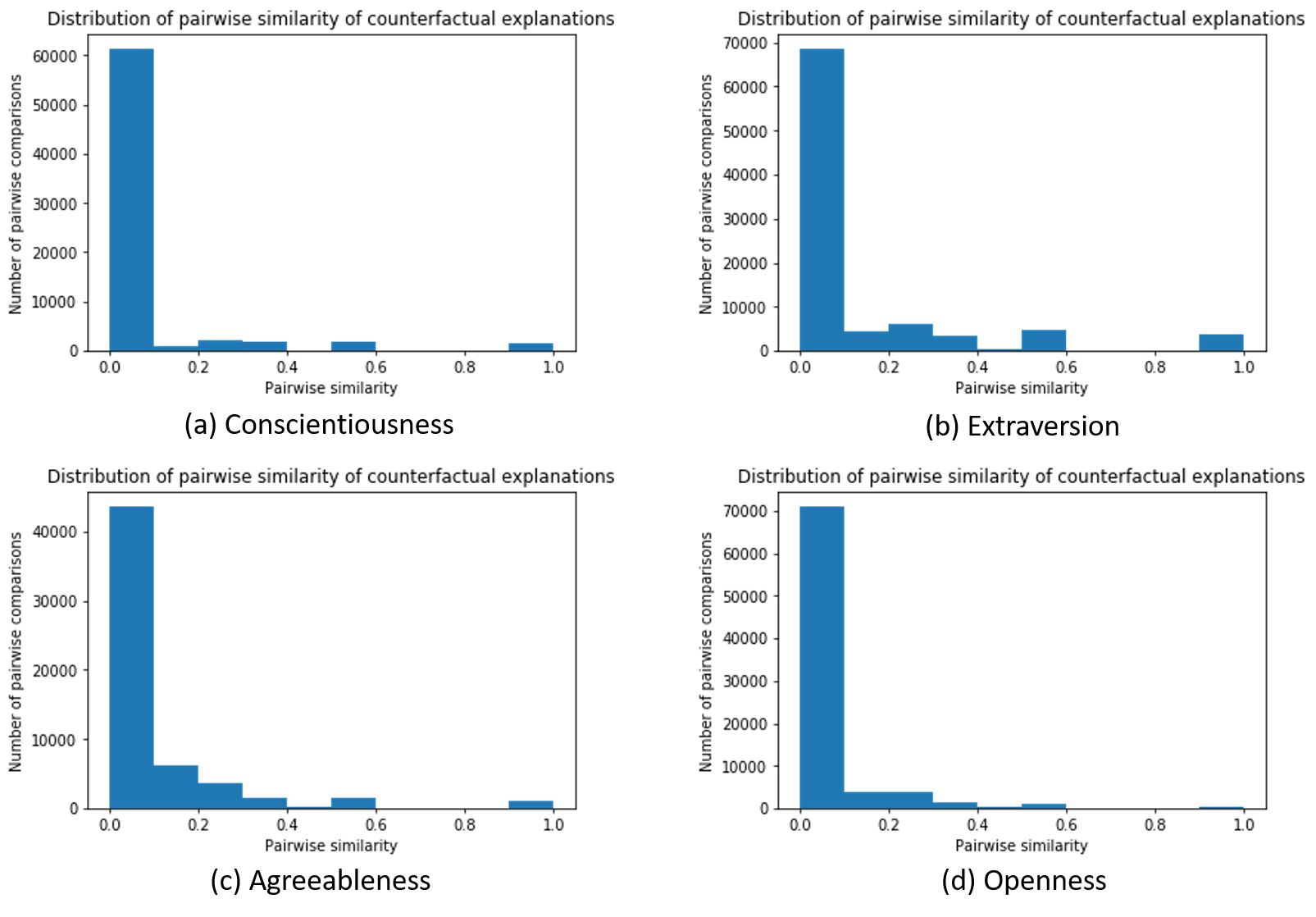}}
	\caption{\textit{Similarity of explanations.} Distribution of pairwise similarity between counterfactual explanations for predicting (a) Conscientiousness, (b) Extraversion, (c) Agreeableness and (d) Openness. A value of $0$ (resp., $1$) indicates no (resp., perfect) overlap.}
	\label{fig:pairwise_others}
\end{figure}

% *** Fig. ***
\begin{figure}[ht]
	\centering
	\scalebox{0.65}{\includegraphics{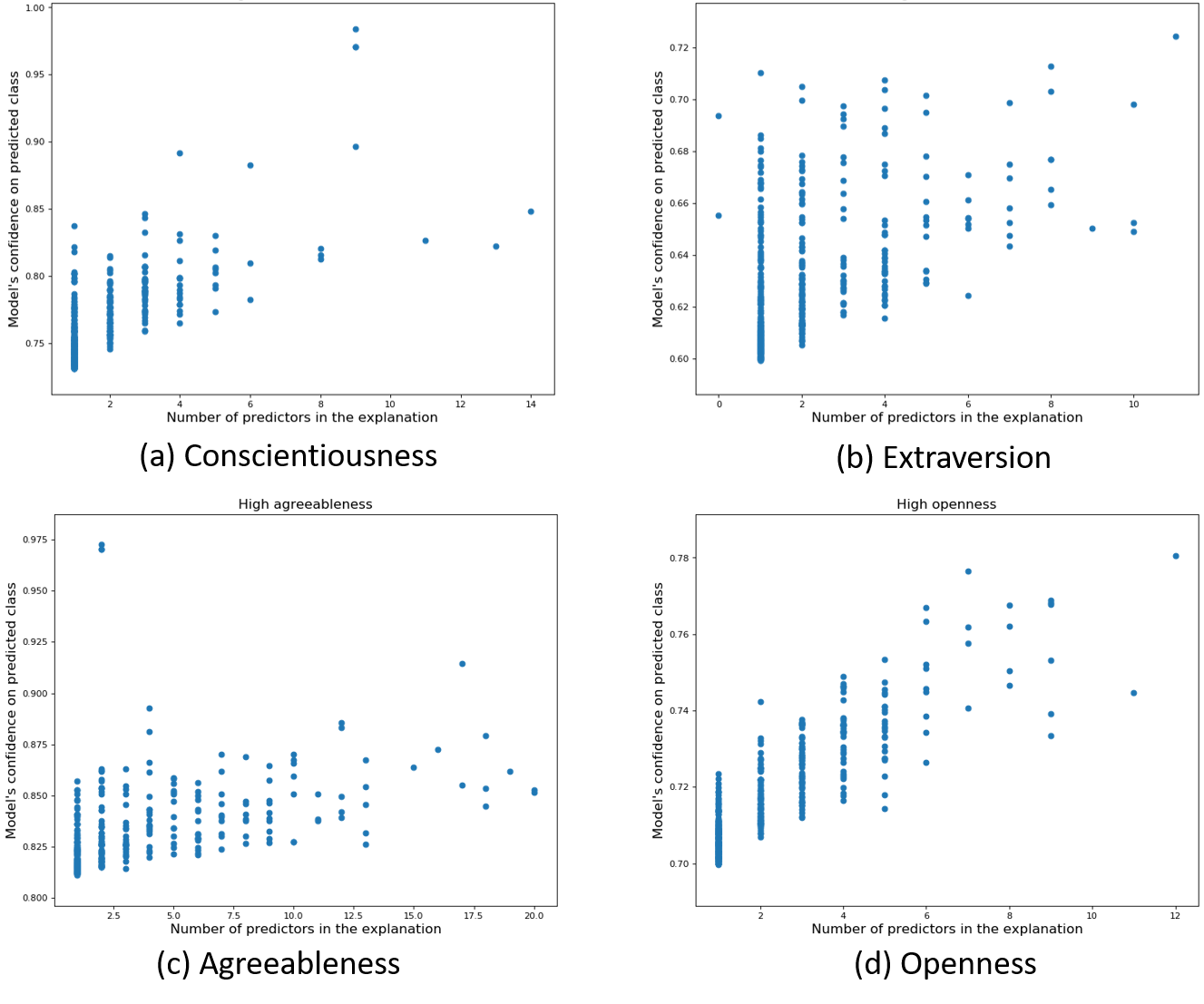}}
	\caption{\textit{Model's classification confidence vs. size of counterfactual explanations.} Model's predicted score vs. number of predictors in the explanation to counterfactually explain the predicted class. The correlations between the confidence scores and explanation sizes are $.72$ (Conscientiousness), $.52$ (Extroversion), $0.44$ (Agreeableness), $.87$ (Openness).}
	\label{fig:ConfvsSize_others}
\end{figure}

\newpage
\bibliographystyle{spmpsci}

\end{document}